\documentclass[11pt]{article}

\usepackage[preprint]{acl}

\usepackage{times}
\usepackage{latexsym}

\usepackage[T1]{fontenc}

\usepackage[utf8]{inputenc}

\usepackage{microtype}

\usepackage{inconsolata}

\usepackage{graphicx}

\usepackage{booktabs}       
\usepackage{multirow}       
\usepackage{siunitx}        

\usepackage{amsmath}
\usepackage{amssymb}

\usepackage{enumitem}

\usepackage{titlesec}
\titlespacing*{\section}{0pt}{1.5ex plus 0.5ex minus 0.2ex}{0.8ex plus 0.2ex}
\titlespacing*{\subsection}{0pt}{1.2ex plus 0.4ex minus 0.2ex}{0.6ex plus 0.2ex}
\titlespacing*{\subsubsection}{0pt}{1ex plus 0.3ex minus 0.1ex}{0.5ex plus 0.1ex}
\titlespacing*{\paragraph}{0pt}{0.8ex plus 0.2ex minus 0.1ex}{0.5em}

\usepackage{algorithm}
\usepackage{algpseudocode}

\usepackage{float}          

\usepackage{listings}

\usepackage{tcolorbox}
\tcbuselibrary{breakable,skins,listings}

\usepackage{caption}
\AtBeginDocument{
  \setlength{\abovedisplayskip}{1pt}
  \setlength{\belowdisplayskip}{1pt}
  \setlength{\abovedisplayshortskip}{2pt}
  \setlength{\belowdisplayshortskip}{2pt}
}

\setlist[itemize]{leftmargin=*, nosep}
\setlist[enumerate]{leftmargin=*, nosep}

\tcbset{
  promptbox/.style={
    enhanced,
    breakable,
    colback=gray!3,
    colframe=gray!60,
    boxrule=0.5pt,
    arc=2pt,
    left=6pt, right=6pt, top=6pt, bottom=6pt,
    fontupper=\small\ttfamily,
    before upper={\parindent0pt\parskip4pt}
  },
  systemprompt/.style={
    enhanced,
    breakable,
    colback=gray!3,
    colframe=gray!70,
    coltitle=black,
    fonttitle=\small\bfseries,
    boxrule=0.5pt,
    arc=2pt,
    left=6pt, right=6pt, top=6pt, bottom=6pt,
    fontupper=\small\ttfamily,
    before upper={\parindent0pt\parskip4pt},
    attach boxed title to top left={yshift=-2mm, xshift=4mm},
    boxed title style={colback=gray!20, colframe=gray!70, arc=1pt, boxrule=0.3pt}
  },
  humanprompt/.style={
    enhanced,
    breakable,
    colback=blue!2,
    colframe=blue!40,
    coltitle=black,
    fonttitle=\small\bfseries,
    boxrule=0.5pt,
    arc=2pt,
    left=6pt, right=6pt, top=6pt, bottom=6pt,
    fontupper=\small\ttfamily,
    before upper={\parindent0pt\parskip4pt},
    attach boxed title to top left={yshift=-2mm, xshift=4mm},
    boxed title style={colback=blue!10, colframe=blue!40, arc=1pt, boxrule=0.3pt}
  },
  assistantprompt/.style={
    enhanced,
    breakable,
    colback=green!2,
    colframe=green!40!black,
    coltitle=black,
    fonttitle=\small\bfseries,
    boxrule=0.5pt,
    arc=2pt,
    left=6pt, right=6pt, top=6pt, bottom=6pt,
    fontupper=\small\ttfamily,
    before upper={\parindent0pt\parskip4pt},
    attach boxed title to top left={yshift=-2mm, xshift=4mm},
    boxed title style={colback=green!10, colframe=green!40!black, arc=1pt, boxrule=0.3pt}
  },
  rubricbox/.style={
    enhanced,
    breakable,
    colback=yellow!3,
    colframe=yellow!50!black,
    coltitle=black,
    fonttitle=\small\bfseries,
    boxrule=0.5pt,
    arc=2pt,
    left=6pt, right=6pt, top=6pt, bottom=6pt,
    fontupper=\small\ttfamily,
    before upper={\parindent0pt\parskip4pt},
    attach boxed title to top left={yshift=-2mm, xshift=4mm},
    boxed title style={colback=yellow!15, colframe=yellow!50!black, arc=1pt, boxrule=0.3pt}
  },
  codebox/.style={
    enhanced,
    breakable,
    colback=gray!3,
    colframe=gray!50,
    coltitle=black,
    fonttitle=\small\bfseries,
    boxrule=0.5pt,
    arc=2pt,
    left=4pt, right=4pt, top=4pt, bottom=4pt,
    fontupper=\scriptsize\ttfamily,
    before upper={\parindent0pt\parskip2pt},
    attach boxed title to top left={yshift=-2mm, xshift=4mm},
    boxed title style={colback=gray!15, colframe=gray!50, arc=1pt, boxrule=0.3pt}
  },
  examplebox/.style={
    enhanced,
    breakable,
    colback=purple!2,
    colframe=purple!40,
    coltitle=black,
    fonttitle=\small\bfseries,
    boxrule=0.5pt,
    arc=2pt,
    left=6pt, right=6pt, top=6pt, bottom=6pt,
    fontupper=\small,
    before upper={\parindent0pt\parskip4pt},
    attach boxed title to top left={yshift=-2mm, xshift=4mm},
    boxed title style={colback=purple!10, colframe=purple!40, arc=1pt, boxrule=0.3pt}
  }
}

\newtcolorbox{promptboxenv}{promptbox}
\newtcolorbox{systempromptbox}[1][System]{systemprompt, title=#1}
\newtcolorbox{userpromptbox}[1][User]{humanprompt, title=#1}
\newtcolorbox{assistantbox}[1][Assistant]{assistantprompt, title=#1}
\newtcolorbox{rubricboxenv}[1][Rubric]{rubricbox, title=#1}
\newtcolorbox{codeboxenv}[1][Code]{codebox, title=#1}
\newtcolorbox{exampleboxenv}[1][Example]{examplebox, title=#1}

\newtcblisting{pythonbox}[1][Python]{
  listing only,
  breakable,
  enhanced,
  colback=gray!3,
  colframe=gray!50,
  coltitle=black,
  fonttitle=\small\bfseries,
  boxrule=0.5pt,
  arc=2pt,
  left=4pt, right=4pt, top=4pt, bottom=4pt,
  attach boxed title to top left={yshift=-2mm, xshift=4mm},
  boxed title style={colback=gray!15, colframe=gray!50, arc=1pt, boxrule=0.3pt},
  title=#1,
  listing options={
    language=Python,
    basicstyle=\scriptsize\ttfamily,
    keywordstyle=\color{blue!70!black},
    commentstyle=\color{green!50!black},
    stringstyle=\color{orange!70!black},
    showstringspaces=false,
    breaklines=true,
    numbers=none
  }
}

\newtcblisting{pythonboxnum}[1][Python]{
  listing only,
  breakable,
  enhanced,
  colback=gray!3,
  colframe=gray!50,
  coltitle=black,
  fonttitle=\small\bfseries,
  boxrule=0.5pt,
  arc=2pt,
  left=4pt, right=4pt, top=4pt, bottom=4pt,
  attach boxed title to top left={yshift=-2mm, xshift=4mm},
  boxed title style={colback=gray!15, colframe=gray!50, arc=1pt, boxrule=0.3pt},
  title=#1,
  listing options={
    language=Python,
    basicstyle=\scriptsize\ttfamily,
    keywordstyle=\color{blue!70!black},
    commentstyle=\color{green!50!black},
    stringstyle=\color{orange!70!black},
    showstringspaces=false,
    breaklines=true,
    numbers=left,
    numberstyle=\tiny\color{gray},
    numbersep=5pt
  }
}

\lstdefinestyle{pythonstyle}{
  language=Python,
  basicstyle=\scriptsize\ttfamily,
  keywordstyle=\color{blue!70!black},
  commentstyle=\color{green!50!black},
  stringstyle=\color{orange!70!black},
  showstringspaces=false,
  breaklines=true,
  frame=none,
  numbers=none,
  aboveskip=2pt,
  belowskip=2pt
}

\lstdefinestyle{pythonstylenum}{
  language=Python,
  basicstyle=\scriptsize\ttfamily,
  keywordstyle=\color{blue!70!black},
  commentstyle=\color{green!50!black},
  stringstyle=\color{orange!70!black},
  showstringspaces=false,
  breaklines=true,
  frame=single,
  framerule=0.4pt,
  rulecolor=\color{gray!50},
  numbers=left,
  numberstyle=\tiny\color{gray},
  numbersep=5pt,
  aboveskip=4pt,
  belowskip=4pt
}

\lstdefinestyle{jsonstyle}{
  basicstyle=\scriptsize\ttfamily,
  stringstyle=\color{orange!70!black},
  showstringspaces=false,
  breaklines=true,
  frame=none,
  numbers=none,
  literate=
    *{:}{{{\color{blue!70!black}:}}}{1}
    {,}{{{\color{blue!70!black},}}}{1}
    {\{}{{{\color{gray!70!black}\{}}}{1}
    {\}}{{{\color{gray!70!black}\}}}}{1}
    {[}{{{\color{gray!70!black}[}}}{1}
    {]}{{{\color{gray!70!black}]}}}{1}
}

\lstdefinestyle{bashstyle}{
  language=bash,
  basicstyle=\scriptsize\ttfamily,
  keywordstyle=\color{blue!70!black},
  commentstyle=\color{green!50!black},
  stringstyle=\color{orange!70!black},
  showstringspaces=false,
  breaklines=true,
  frame=none,
  numbers=none,
  aboveskip=2pt,
  belowskip=2pt
}

\newcommand{\eg}{\textit{e.g.}}

\DeclareMathOperator*{\argmax}{arg\,max}

\newcommand{\best}[1]{\textbf{#1}}

\title{LLM-Augmented Changepoint Detection: \\ A Framework for Ensemble Detection and Automated Explanation}

\author{
\textbf{Fabian Lukassen}$^{1}$,
\textbf{Christoph Weisser}$^{3}$,
\textbf{Michael Schlee}$^{1}$,
\textbf{Manish Kumar}$^{2}$,
\textbf{Anton Thielmann}$^{1}$\thanks{Work completed while at TU Clausthal.},
\\
\textbf{Benjamin Saefken}$^{2}$,
\textbf{Alexander Silbersdorff}$^{1}$,
\textbf{Thomas Kneib}$^{1}$ \\
$^{1}$University of G\"ottingen \quad $^{2}$TU Clausthal \quad $^{3}$ Hochschule Bielefeld  \\
\texttt{fabian.lukassen@stud.uni-goettingen.de}
\texttt{\{tkneib, asilbersdorff, michael.schlee\}@uni-goettingen.de} \\
\texttt{christoph.weisser@hsbi.de}
\texttt{\{benjamin.saefken, manish.kumar.2\}@tu-clausthal.de} \\
\texttt{antonthielmann@t-online.de}
}

\begin{document}
\maketitle

\begin{abstract}
This paper introduces a novel changepoint detection framework that combines ensemble statistical methods with Large Language Models (LLMs) to enhance both detection accuracy and the interpretability of regime changes in time series data. Two critical limitations in the field are addressed. First, individual detection methods exhibit complementary strengths and weaknesses depending on data characteristics, making method selection non-trivial and prone to suboptimal results. Second, automated, contextual explanations for detected changes are largely absent. The proposed ensemble method aggregates results from ten distinct changepoint detection algorithms, achieving superior performance and robustness compared to individual methods. Additionally, an LLM-powered explanation pipeline automatically generates contextual narratives, linking detected changepoints to potential real-world historical events. For private or domain-specific data, a Retrieval-Augmented Generation (RAG) solution enables explanations grounded in user-provided documents. The open source Python framework demonstrates practical utility in diverse domains, including finance, political science, and environmental science, transforming raw statistical output into actionable insights for analysts and decision-makers.\footnote{Package: \url{https://anonymous.4open.science/r/Ensemble_Changepoint_Detection-8BD1/}}
\end{abstract}

\section{Introduction}

\begin{figure}[t]
\centering
\includegraphics[width=\columnwidth]{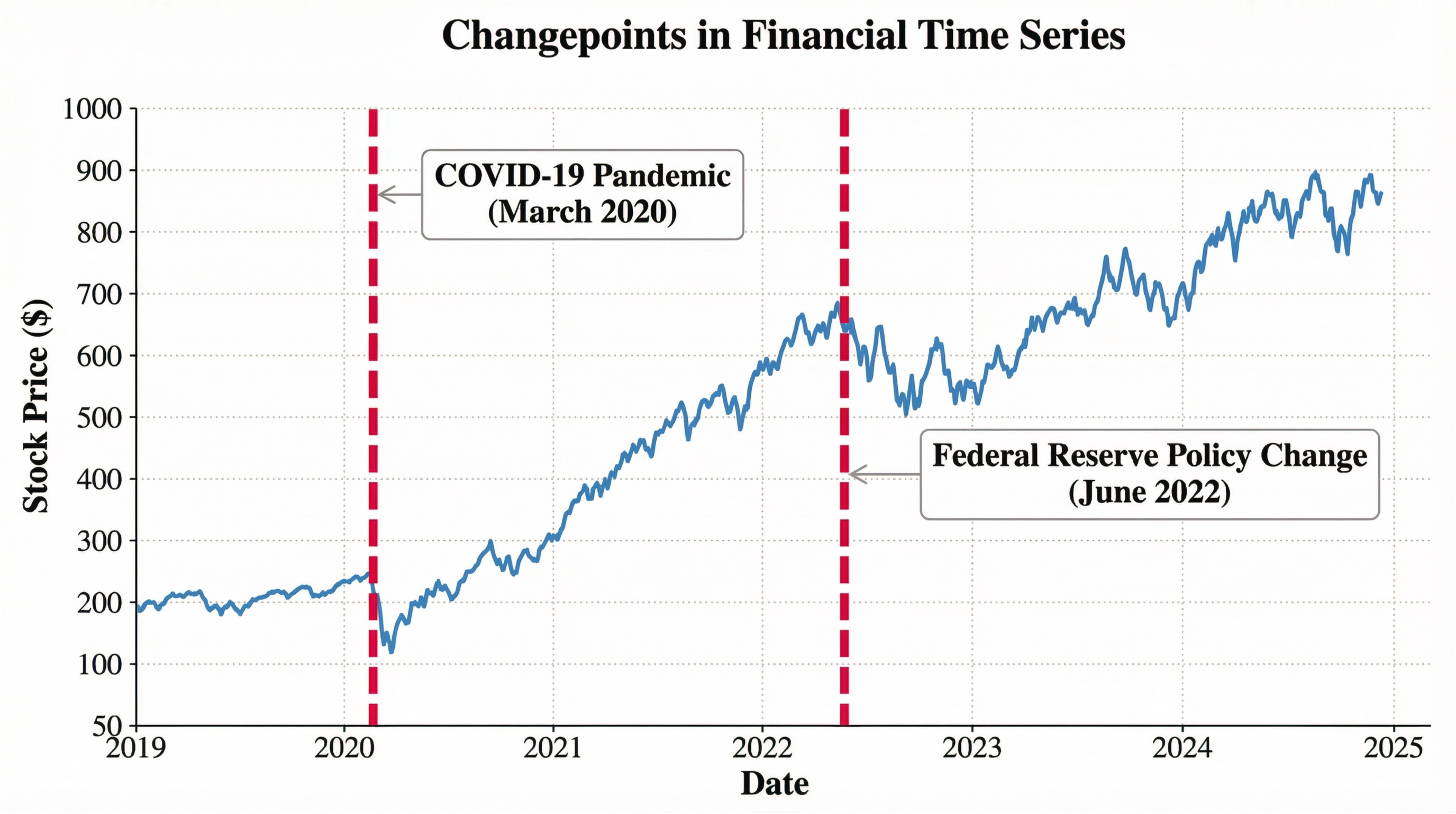}
\caption{Structural break analysis in financial time series. Detected changepoints (red dashed lines) correspond to major real-world events. This illustrates our framework's core challenge: automatically linking statistical anomalies to their historical causes.}
\label{fig:structural_break}
\end{figure}

Changepoint detection represents a fundamental challenge in time series analysis, with applications spanning economics, environmental science, public health, and numerous other domains \cite{truong2020ruptures}. In general, changepoint detection involves identifying moments when the statistical properties of a time series undergo an abrupt change, indicating regime changes, policy interventions, market crashes, or other significant events. Although traditional statistical methods have shown effectiveness in detecting the temporal location of these changes, they face two persistent limitations that our work addresses.

First, no single detection algorithm performs optimally across all types of time series and changepoint characteristics. Following the spirit of the \textit{no free lunch} theorem \cite{wolpert1997no}, methods such as CUSUM \cite{page1954cusum} are excellent at detecting mean shifts in stationary series, while others like the Bai-Perron test \cite{bai2003computation} are superior for multiple breaks in regression settings, and PELT \cite{killick2012pelt} handles complex change patterns efficiently. This diversity of strengths and weaknesses necessitates careful method selection based on data characteristics, a process that requires substantial expertise and often involves trial-and-error approaches.

Second, traditional methods provide only statistical evidence of the changepoints without contextual explanations of why these changes occurred. Analysts must manually investigate historical records, news archives, and domain-specific knowledge to understand the underlying causes \cite{merelo2024pivotal}. This interpretive burden is time-consuming, subjective, and increasingly impractical as data volumes grow exponentially across industries and research domains.

The emergence of Large Language Models (LLMs) presents an unprecedented opportunity to address this interpretive challenge. These systems possess a vast knowledge base that encompasses historical events, economic developments, and domain-specific information, combined with sophisticated natural language generation capabilities. However, effectively leveraging LLMs for changepoint analysis requires careful integration with statistical methods and appropriate handling of both public and private data contexts.

Our research introduces the \textbf{LLM-Augmented Changepoint Detection} framework, which makes three primary contributions:
\begin{enumerate}
    \item \textbf{Ensemble Detection Method:} An ensemble approach that combines ten distinct changepoint detection algorithms through a consensus-based voting mechanism. This method includes spatial clustering of nearby detections and adaptive confidence scoring, achieving superior performance compared to single algorithm approaches on diverse real-world datasets.
    \item \textbf{Automated LLM Explanations:} An explanation system that automatically generates contextual narratives for detected changepoints. The system prompts LLMs with temporal context and data characteristics to identify plausible historical events that coincide with statistical anomalies, transforming abstract mathematical findings into actionable insights.
    \item \textbf{RAG-Enhanced Private Data Support:} For organizations working with proprietary or sensitive time series data (absent from LLM training corpora), we provide a Retrieval-Augmented Generation (RAG) solution that enables explanations based on user-provided context documents. This maintains data privacy while enabling factual, domain-specific explanations that would otherwise be impossible.
\end{enumerate}

The framework is implemented as an open-source Python package with a modular architecture that supports flexible integration of detection methods, multiple LLM providers, and custom domain adaptations. Our approach, depicted in Figure \ref{fig:workflow}, demonstrates practical utility in diverse applications, from explaining stock market volatility shifts in finance to contextualizing conflict intensity changes in political science.

\section{Related Work}

Our work is situated at the intersection of automated statistical analysis and natural language interpretation. This section positions our work within the broader context of ensemble methods in changepoint detection, LLMs in time series analysis, and automated explanation systems.

\subsection{Ensemble Changepoint Detection}

Although the literature on individual changepoint detection algorithms is extensive \cite{aminikhanghahi2017survey}, ensemble-based methodologies have recently emerged as a means to enhance robustness and reliability. Several principled ensemble frameworks have been proposed, each implementing distinct aggregation strategies and exhibiting specific methodological trade-offs.

\citet{katser2021cpde} proposed an unsupervised framework (CPDE) that aggregates cost functions from multiple methods before performing a search. \citet{zhao2019beast} introduced BEAST, a Bayesian model averaging approach that provides probabilistic estimates of changepoints. Other techniques include data-centric ensembles like Ensemble Binary Segmentation (EBS), which applies a single algorithm to multiple data subsamples \cite{korkas2022ebs}, and feature-centric ensembles like PCA-uCPD, which runs detection on different principal components of multivariate data \cite{qin2025pcaucpd}. Machine learning-based approaches such as ChangeForest \cite{londschien2023changeforest} and deep learning methods such as WWAggr \cite{stepikin2025wwaggr} have also been developed.

Our work contributes to this space by creating a practical, flexible ensemble that aggregates the final outputs of diverse, complete algorithms rather than intermediate scores or features. Methods such as CPDE \cite{katser2021cpde} that aggregate cost functions and WWAggr \cite{stepikin2025wwaggr} that leverage complex Wasserstein distance calculations follow elaborate strategies. In contrast, our approach prioritizes interpretability and simplicity.

We introduce a spatial clustering step that is unique in its handling of near-coincident detections from different methods, combined with a transparent voting mechanism that allows analysts to understand exactly which methods contributed to each detected changepoint. This design philosophy makes our ensemble more accessible to practitioners while maintaining robust performance.

\subsection{Event Attribution and Automated Explanation}

Manual attribution of statistical findings to real-world events is a long-standing practice in analytics. Recently, \citet{merelo2024pivotal} demonstrated a methodology for applying changepoint analysis to historical time series and manually validating detected points against known historical events. Our work automates this process using LLMs.

The idea of using LLMs for explanation is not new, but its application to time series is nascent. \citet{zhang2024laserbeam} proposed LASER-BEAM to generate narratives explaining \textit{forecasts} of stock market volatility, a forward-looking task. In contrast, our work focuses on the post-hoc explanation of \textit{detected} historical changepoints. Other related works like iPrompt \cite{singh2022iprompt} and TCube \cite{sharma2021tcube} focus on generating explanations for general data patterns or narrating time series, but do not specifically target the causal attribution of discrete changepoints.

\subsection{LLMs for Time Series Analysis}

Recent research has explored the capabilities of LLMs for various time series tasks. Models like AXIS \cite{lan2025axis} have been developed for explainable anomaly detection, which is related but distinct from changepoint detection. Anomalies are usually single-point deviations, whereas changepoints are persistent shifts in statistical properties. TsLLM \cite{parker2025tsllm} and other similar models aim to create a unified foundation model for general time series tasks such as forecasting and classification, but do not specialize in the attribution of historical events of structural breaks.

\subsection{Retrieval-Augmented Generation (RAG)}

To address the limitations of internal knowledge of LLMs, especially for recent or private data, we incorporate Retrieval-Augmented Generation (RAG) \cite{lewis2020rag, reuter2025gptopic}. RAG is well-established, but its use for explaining time series is novel to the knowledge of the authors. RAAD-LLM \cite{russellgilbert2025raadllm} applied a similar concept to anomaly detection in system logs, but our work is the first to use RAG to provide contextually rich and factually grounded explanations for changepoints in any arbitrary time series by retrieving from a user-provided corpus of documents.

\subsection{Research Gaps and Our Contribution}

We identify a clear gap in the literature: no existing framework integrates robust, ensemble-based changepoint detection with automated, LLM-powered historical event attribution. Our primary contribution is to bridge this gap, creating a seamless pipeline from statistical detection to human-readable insight. We automate the manual attribution process demonstrated by \citet{merelo2024pivotal} and adapt the explanatory power of LLMs from forecasting \cite{zhang2024laserbeam} and anomaly detection \cite{lan2025axis} to the specific post-hoc challenge of explaining the changepoint.

\section{Problem Statement}

Let a univariate time series be represented as an ordered sequence of $n$ observations $X = \{x_t\}_{t=1}^n$, where $x_t \in \mathbb{R}$ is the observation at time $t$. A changepoint is a time index $\tau \in \{2, \dots, n-1\}$ where the statistical properties (\eg, mean, variance, or distribution) of the sequence before and after $\tau$ are significantly different. The core task of changepoint detection is to estimate the set of $m$ changepoint locations $\mathcal{T} = \{\tau_1, \dots, \tau_m\}$.

The $m$ changepoints divide the time series into $m+1$ contiguous segments $S_0, \dots, S_m$. With boundary indices $\tau_0 = 1$ and $\tau_{m+1} = n+1$ (ensuring coverage from $x_1$ to $x_n$):

$$ X = \bigsqcup_{j=0}^{m} S_j, \quad S_j = \{x_{\tau_j}, \dots, x_{\tau_{j+1}-1}\} $$

\noindent where the disjoint union $\bigsqcup$ preserves the temporal ordering of the segments.

Given a time series $X$ and an optional corpus of private documents $\mathcal{D}$, the goal then is to produce a set of tuples $\mathcal{R} = \{(\hat{\tau}_i, c_i, E_i)\}_{i=1}^{m}$, where: (1) $\hat{\tau}_i$ is the estimated location of the $i$-th detected changepoint; (2) $c_i \in [0, 1]$ is the confidence score associated with the detection of $\hat{\tau}_i$; and (3) $E_i$ is a natural language text that provides a plausible and contextually relevant historical explanation for the change observed at $\hat{\tau}_i$, grounded in public knowledge or the private corpus $\mathcal{D}$.

\section{Methodology}

\begin{figure*}[!ht]
\centering
\includegraphics[width=0.65\textwidth]{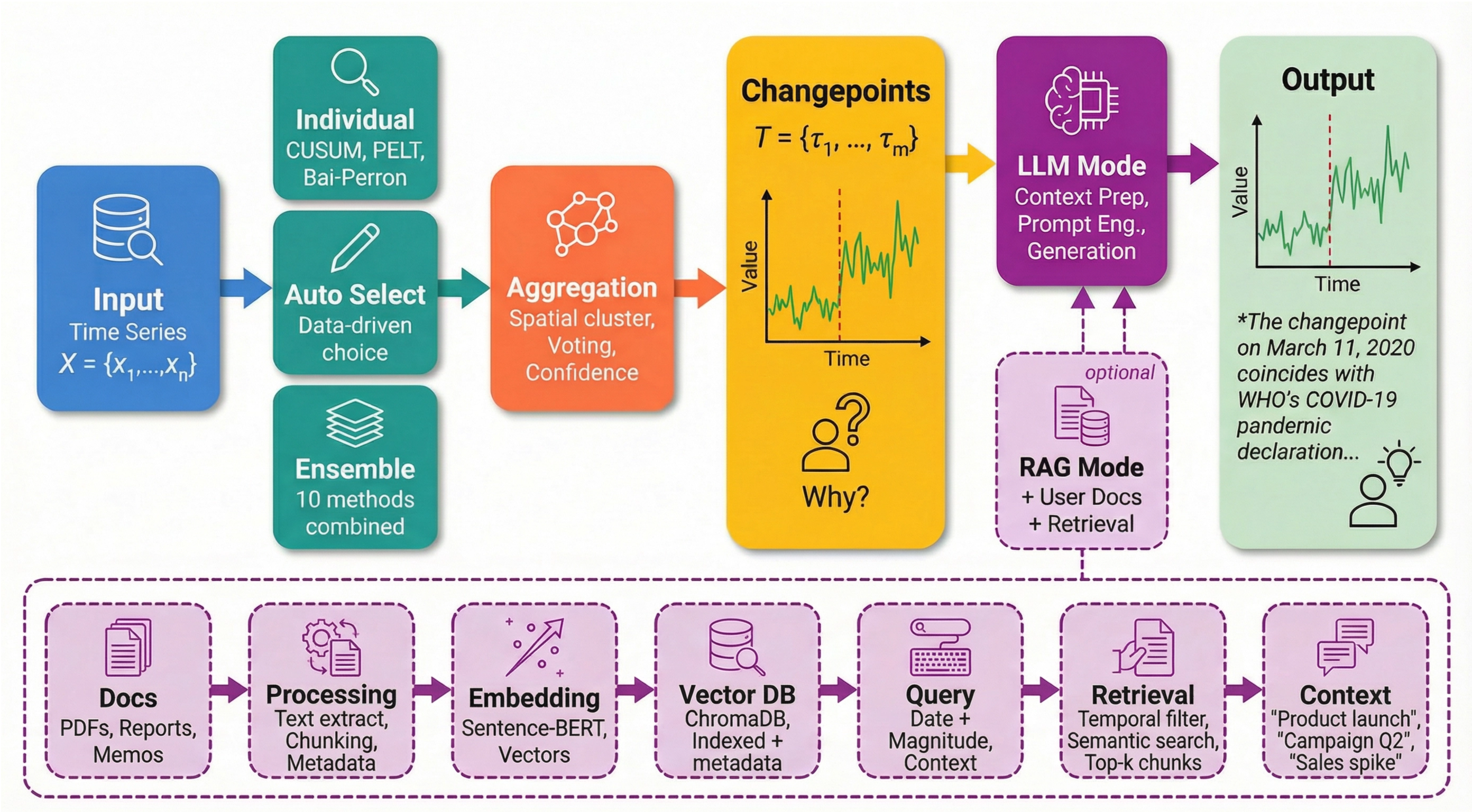}
\caption{Complete workflow of the LLM-augmented changepoint detection framework. The system processes time series data through three main stages: (1) changepoint detection (2) optional RAG integration for private data, retrieving relevant documents using hybrid semantic-temporal search; and (3) LLM-based causal attribution, which identifies plausible events underlying the detected changepoints.}
\label{fig:workflow}
\end{figure*}

Beyond the core changepoint detection and explanation pipeline detailed below, the Python package provides additional capabilities: general time series diagnostics (\eg, stationarity tests, trend and seasonality analysis) with LLM-powered natural language explanations, automated report generation in multiple formats (HTML, PDF, Markdown), an interactive web interface for no-code analysis, and a Jupyter widget. These features, along with usage examples for LLM-based explanations, event attribution via common API providers, and RAG document management, are demonstrated in Appendix~\ref{app:additional_capabilities}.

This section focuses on our three main contributions: (1) changepoint detection, (2) LLM-based event attribution, and (3) an optional RAG pipeline for private data. The workflow is illustrated in Figure~\ref{fig:workflow}.

\subsection{Changepoint Detection Strategies}

We provide three detection strategies to accommodate different use cases and expertise levels.

\subsubsection{Individual Method Selection}

Users with domain expertise can select from ten distinct changepoint detection algorithms, grouped by their underlying approach: \textbf{Statistical Tests} such as CUSUM \cite{page1954cusum} for the detection of mean shifts; \textbf{Segmentation Methods} including PELT \cite{killick2012pelt}, Binary Segmentation \cite{scott1974binary}, Bottom-Up \cite{keogh2001bottomup}, and Window-based methods for optimal partitioning; \textbf{Kernel-Based Methods} like Kernel CPD \cite{harchaoui2007kernel} for distribution changes; and \textbf{Bayesian Methods} such as Bayesian Online CPD \cite{adams2007bayesian} for probabilistic estimates. Each method produces changepoint locations. Note that each detection method applies method-specific preprocessing to the data (\eg, stationarity transformations, missing value handling, outlier treatment) and computes confidence scores for detected changepoints using method-specific formulations; complete details for each are provided in Appendix~\ref{app:cpd_methods}.

\subsubsection{Automatic Method Selection}

\begin{algorithm}[H]
\small
\caption{Automatic Method Selection}
\label{alg:auto_select}
\begin{algorithmic}[1]
\Require Time series data $X = \{x_1, \ldots, x_n\}$
\Ensure Selected detection method $m^*$
\State \textbf{// Phase 1: Data Profiling}
\State $n \gets |X|$
\State $\nu \gets \sigma(X) / |\mu(X)|$ \Comment{Coefficient of variation}
\State $\rho \gets |\text{corr}(X, \{1, \ldots, n\})|$ \Comment{Trend strength}
\State $s \gets \text{ADF\_test}(X).\text{pvalue}$ \Comment{Stationarity}
\State $o \gets \text{outlier\_ratio}(X)$ \Comment{IQR-based outliers}
\State $\lambda \gets \text{seasonality\_score}(X)$ \Comment{Pattern check}
\State
\State \textbf{// Phase 2: Method Filtering}
\State $\mathcal{M}_{\text{all}} \gets \{\text{bai\_perron}, \text{cusum}, \text{pelt}, \ldots\}$
\State $\mathcal{M}_{\text{valid}} \gets \emptyset$
\For{each method $m \in \mathcal{M}_{\text{all}}$}
    \If{$m$ satisfies data size requirements for $n$}
        \State $\mathcal{M}_{\text{valid}} \gets \mathcal{M}_{\text{valid}} \cup \{m\}$
    \EndIf
\EndFor
\State
\State \textbf{// Phase 3: Multi-Criteria Scoring}
\For{each method $m \in \mathcal{M}_{\text{valid}}$}
    \State $\text{score}(m) \gets \sum_{i=1}^{7} f_i(m, \text{char}_i)$
\EndFor
\State
\State \textbf{// Phase 4: Selection}
\State $m^* \gets \argmax_{m \in \mathcal{M}_{\text{valid}}} \text{score}(m)$
\State \Return $m^*$
\end{algorithmic}
\end{algorithm}

For users without expertise in changepoint detection, we provide an automatic method selection strategy based on a \textbf{Multi-Criteria Decision Matrix}. This approach systematically evaluates each detection method in seven dimensions of the data characteristics, selecting the method most suited to the specific properties of the input time series.

\paragraph{Data Profiling}
The system first extracts six key characteristics from the time series $X = \{x_1, \ldots, x_n\}$:

\begin{enumerate}
    \item \textbf{Sample Size} ($n$): Total number of observations.
    \item \textbf{Noise Level} ($\nu$): Coefficient of variation, $\nu = \sigma/(|\mu| + \epsilon)$, where $\sigma$ and $\mu$ are the standard deviation and mean of $X$, and $\epsilon = 10^{-8}$ prevents division by zero.
    \item \textbf{Trend Strength} ($\rho$): Absolute Pearson correlation between $X$ and the deterministic time index $t = \{1, \ldots, n\}$:
    $$\rho = |\text{corr}(X, t)|$$
 This metric may underestimate non-linear trends and can be affected by seasonality.
    \item \textbf{Stationarity} ($s$): Augmented Dickey-Fuller (ADF) test p-value. Lower values ($s < 0.05$) provide evidence for stationarity; higher values indicate insufficient evidence to reject the unit root hypothesis.
    \item \textbf{Outlier Ratio} ($o$): Proportion of detrended observations outside the interquartile range (IQR) bounds:
    \begin{equation*}
    \begin{split}
    o = \frac{1}{n} \sum_{i=1}^{n} \mathbb{1}[r_i < Q_1 - 1.5 \cdot \text{IQR} \\
    \text{ or } r_i > Q_3 + 1.5 \cdot \text{IQR}]
    \end{split}
    \end{equation*}
    where $r_i$ are detrended residuals and $Q_1, Q_3$ are the first and third quartiles of $r$.
    \item \textbf{Seasonality} ($\lambda$): Maximum absolute autocorrelation across candidate lags:
    $$\lambda = \max_{k \in \{7, 12, 24, 30, 365\}} |\text{ACF}(k)|$$
    where ACF$(k)$ is the autocorrelation function at lag $k$, covering weekly, monthly, and yearly patterns.
\end{enumerate}

\paragraph{Method Filtering and Scoring}
Before scoring, we filter candidate methods based on their minimum data requirements (detailed in Appendix~\ref{app:auto_selection_details}). Each remaining method $m$ is then evaluated across seven scoring criteria $\{f_1, \ldots, f_7\}$, where each $f_i: (\text{method}, \text{characteristic}) \rightarrow [0, 1]$ quantifies suitability for: (1) sample size, (2) noise tolerance, (3) trend handling, (4) seasonality handling, (5) computational efficiency, (6) stationarity handling, and (7) outlier robustness. Note that some criteria use the same characteristic (e.g., both sample size suitability and computational efficiency depend on $n$). The total score is:

$$\text{score}(m) = \sum_{i=1}^{7} f_i(m, \text{char}_i)$$

\noindent where $\text{char}_i \in \{n, \nu, \rho, s, o, \lambda\}$ maps each criterion to its relevant characteristic. The method with maximum score is selected: $m^* = \arg\max_{m \in \mathcal{M}_{\text{valid}}} \text{score}(m)$. This approach automates expert method selection while remaining interpretable: users can inspect which data characteristics drove the selection. The procedure is formalized in Algorithm~\ref{alg:auto_select}, with complete scoring matrices in Appendix~\ref{app:auto_selection_details}.

\subsubsection{Ensemble Method}

For maximum robustness, the ensemble method runs all ten algorithms independently and aggregates their output. The aggregation proceeds in three steps:

\paragraph{Step 1: Spatial Clustering}
Raw detections from different methods are often temporally proximate but not identical. Let $\mathcal{D}_j$ be the set of changepoint time indices detected by method $j$, where $j \in \{1, \dots, 10\}$. The total set of raw, unrefined detections is given by the union:
$$\mathcal{D} = \bigcup_{j=1}^{10} \mathcal{D}_j$$

We apply agglomerative clustering to the set $\mathcal{D}$ using a maximum temporal proximity threshold $\epsilon$, a hyperparameter controlling cluster granularity (i.e., the maximum distance allowed between elements to be in the same cluster). This process yields a set of $m$ clusters $\mathcal{C}$, where each cluster $C_i$ represents a unified consensus changepoint:
$$\mathcal{C} = \text{Cluster}(\mathcal{D}, \epsilon) = \{C_1, \dots, C_m\}$$

\noindent In our implementation, we set $\epsilon = \min(5, \max(2, n/40))$; the scaling factor of 40 was determined empirically to balance sensitivity and specificity across diverse time series lengths.

\paragraph{Step 2: Voting}
For each cluster $C_i$, we count the number of unique methods that contributed:

$$v_i = |\{j : \exists t \in C_i \text{ detected by method } j\}|$$

This vote count $v_i$ quantifies the strength of the consensus for the changepoint.

\paragraph{Step 3: Thresholding and Aggregation}
A cluster is confirmed as a final changepoint if $v_i \geq v_{\text{min}}$ (e.g. $v_{\text{min}} = 5$). The final changepoint location and confidence are:

$$\hat{t}_i = \frac{\sum_{t \in C_i} c(t) \cdot t}{\sum_{t \in C_i} c(t)}, \quad \hat{c}_i = \frac{1}{|C_i|} \sum_{t \in C_i} c(t)$$

\noindent where $c(t)$ is the confidence score for detection $t$. This provides a confidence-weighted temporal estimate.

This approach prioritizes \textbf{interpretability}: analysts can inspect which methods voted for each changepoint, understand the consensus level, and adjust $v_{\text{min}}$ based on their risk tolerance. Unlike complex aggregation schemes (e.g., Wasserstein barycenters \cite{stepikin2025wwaggr}), our method is transparent. The complete ensemble procedure is formalized in Algorithm~\ref{alg:ensemble}.

\begin{algorithm}[t]
\small
\caption{Ensemble Changepoint Detection}
\label{alg:ensemble}
\begin{algorithmic}[1]
\Require Time series $X$ with $n$ observations, min votes $v_{\min}$
\Ensure Changepoints $\mathcal{T} = \{(\hat{t}_i, \hat{c}_i)\}$
\State Run all 10 methods on $X$
\State $\mathcal{D} \gets \{(t, c, m) : \text{method } m \text{ detected break at } t\}$
\State \textbf{// Spatial Clustering}
\State $\epsilon \gets \min(5, \max(2, n/40))$ \Comment{Adaptive tolerance}
\State $\{C_1, \ldots, C_m\} \gets \text{AgglomerativeCluster}(\mathcal{D}, \epsilon)$
\State \textbf{// Voting}
\For{each cluster $C_i$}
    \State $v_i \gets |\{\text{unique methods in } C_i\}|$
\EndFor
\State \textbf{// Aggregation}
\State $\mathcal{T} \gets \emptyset$
\For{each $C_i$ with $v_i \geq v_{\min}$}
    \State $\text{conf\_sum} \gets \sum_{(t,c) \in C_i} c$
    \If{$\text{conf\_sum} = 0$}
        \State $\hat{t}_i \gets \frac{1}{|C_i|} \sum_{(t,c) \in C_i} t$
    \Else
        \State $\hat{t}_i \gets \frac{\sum_{(t,c) \in C_i} c \cdot t}{\text{conf\_sum}}$
    \EndIf
    \State $\hat{c}_i \gets \frac{1}{|C_i|} \sum_{(t,c) \in C_i} c$
    \State $\mathcal{T} \gets \mathcal{T} \cup \{(\hat{t}_i, \hat{c}_i)\}$
\EndFor
\State \Return $\mathcal{T}$
\end{algorithmic}
\end{algorithm}

\subsection{LLM Integration for Event Attribution}

Once a set of high-confidence changepoints is identified, the next stage is to generate a historical explanation for each one. The explanation process operates in two modes:

\paragraph{Standard Mode (Public Knowledge).} The LLM directly generates explanations based on its training data: (1) \textbf{Context Construction} constructs a prompt containing the changepoint date, confidence, magnitude, direction, statistical summaries before and after the break, and data description; (2) \textbf{LLM Query} sends the prompt to an LLM (we support OpenAI GPT-4, Anthropic Claude, Azure OpenAI, and local models), which leverages its internal knowledge of historical events; and (3) \textbf{Explanation Generation} produces a narrative linking the statistical change to plausible real-world events.

\paragraph{RAG Mode (Private Data).} For proprietary or domain-specific data, a RAG pipeline retrieves relevant context from user-provided documents before querying the LLM.

We employ prompt engineering strategies to improve the quality of the explanation: requiring the LLM to consider temporal proximity, assess causality, and acknowledge uncertainty when appropriate. The complete prompt templates are in Appendix~\ref{app:llm_prompts}.

\subsection{RAG for Private Data}

The Retrieval-Augmented Generation (RAG) pipeline is shown in Figure \ref{fig:workflow}. This allows the system to generate explanations based on a private corpus of documents provided by the user.

Our RAG implementation uses the following technologies:

\paragraph{Embedding Model:}
We use \textbf{Sentence-Transformers} \cite{reimers2019sentencebert} with the \texttt{all-MiniLM-L6-v2} model (384 dimensions, 80MB) as the default. For higher quality, users can opt for \texttt{all-mpnet-base-v2} (768 dimensions, 420MB). Given a document chunk $d$, the embedding function is:
$$\mathbf{e}_d = \text{Normalize}(\text{SentenceTransformer}(d))$$
Normalization ensures unit length for cosine similarity.

\paragraph{Vector Store:}
We use \textbf{ChromaDB} \cite{chromadb2023} as the persistent vector database. Documents are stored with metadata (date, title, type) for hybrid search.

\paragraph{Retrieval Process}
For a changepoint at time $t_i$ with description $q$:

\begin{enumerate}
    \item \textbf{Temporal Filtering:} Define a window $[t_i - \Delta, t_i + \Delta]$ (e.g. $\Delta = 30$ days). \item \textbf{Semantic Search:} Compute query embedding $\mathbf{e}_q$ and retrieve top-$k$ documents by cosine similarity:
    $$\text{sim}(\mathbf{e}_q, \mathbf{e}_d) = \mathbf{e}_q \cdot \mathbf{e}_d$$
    \item \textbf{Hybrid Ranking:} Combine semantic similarity with temporal proximity:
    \begin{equation*}
    \begin{aligned}
    \text{score}(d) &= \alpha \cdot \text{sim}(\mathbf{e}_q, \mathbf{e}_d) \\
    &\quad + (1-\alpha) \cdot \text{temporal}(d, t_i)
    \end{aligned}
    \end{equation*}
    where $\alpha \in [0,1]$ is a hyperparameter balancing semantic and temporal relevance; we use $\alpha = 0.7$ as the default, determined empirically to prioritize semantic similarity while still accounting for temporal proximity. The temporal relevance function is
    \begin{equation*}
    \text{temporal}(d, t_i) = \max\left(0, 1 - \frac{|\text{date}(d) - t_i|}{\Delta}\right),
    \end{equation*}
    which assigns higher scores to documents closer to the changepoint date.
    \item \textbf{Augmented Prompt:} Prepend retrieved documents to the LLM prompt.
\end{enumerate}

\section{Evaluation}

\subsection{Ensemble Method}

We demonstrate the capabilities of the package by comparing the ensemble method against auto-selection, which in turn uses individual detection methods.

\paragraph{Datasets.} We curate seven benchmark datasets from the Turing Change Point Dataset and other public sources, each with a single, historically documented structural break whose cause is unambiguous (Table~\ref{tab:datasets}). These datasets span diverse domains and vary in length (21--468 observations), frequency (annual to monthly), and change characteristics (mean shifts, trend changes, variance changes).

\begin{table}[h]
\centering
\small
\begin{tabular*}{\columnwidth}{@{\extracolsep{\fill}}llcl@{}}
\toprule
\textbf{Dataset} & \textbf{Domain} & \textbf{N} & \textbf{Event} \\
\midrule
Nile & Hydrology & 100 & Aswan Dam (1898) \\
Seatbelts & Policy & 108 & UK Law (1983) \\
LGA & Aviation & 468 & 9/11 (2001) \\
Ireland Debt & Economics & 21 & Banking Crisis (2009) \\
Ozone & Environment & 54 & Peak Depletion (1993) \\
Robocalls & Telecom & 53 & FCC Ruling (2018) \\
Japan Nuclear & Energy & 40 & Fukushima (2011) \\
\bottomrule
\end{tabular*}
\caption{Benchmark datasets with ground truth structural breaks.}
\label{tab:datasets}
\end{table}

The datasets represent different types of real-world changepoints: \textbf{Nile} (annual river flow, 1871--1970) shows a mean shift when the Aswan Low Dam construction began in 1898 \cite{cobb1978nile}; \textbf{Seatbelts} (monthly UK road casualties, 1976--1984) exhibits a sudden drop following the 1983 compulsory seatbelt law \cite{harvey1986seatbelt}; \textbf{LGA} (monthly LaGuardia Airport passengers, 1977--2015) captures the immediate and sustained impact of the September 11 attacks on air travel \cite{ito2005airline911}; \textbf{Ireland Debt} (annual debt-to-GDP ratio, 2000--2020) shows the dramatic surge following the 2008 banking crisis and subsequent bailout \cite{lane2011irishcrisis}; \textbf{Ozone} (annual Antarctic ozone measurements, 1961--2014) marks the reversal point when Montreal Protocol effects began \cite{solomon2016ozone}; \textbf{Robocalls} (monthly US call volume, 2015--2019) increases sharply after a 2018 federal court ruling loosened FCC restrictions; and \textbf{Japan Nuclear} (annual nuclear share of electricity, 1985--2024) drops precipitously after the 2011 Fukushima disaster led to reactor shutdowns \cite{hayashi2013fukushima}. Most datasets are sourced from the Turing Change Point Dataset \cite{van2020evaluation}.

\paragraph{LLMs.} For event attribution, we query three LLMs spanning different scales and architectures: \textbf{Llama-3.1-8B} \cite{grattafiori2024llama}, a small open-source model run locally; \textbf{GPT-4o} \cite{openai2024gpt4o}, a large commercial model; and \textbf{DeepSeek-R1} \cite{guo2025deepseekr1}, a 671B-parameter Mixture-of-Experts model with 37B active parameters, optimized for reasoning via reinforcement learning. This selection examines how model scale and reasoning specialization affect explanation quality.

\paragraph{Protocol.} Detection is considered correct if the predicted changepoint falls within $\pm$3 data points of the ground truth index. For evaluating LLM-generated explanations, we employ the LLM-as-a-judge framework \cite{zheng2023llmjudge}. Specifically, we use Kimi K2 \cite{kimik2} as the judge. The judge receives the LLM's explanation along with the ground truth event cause and is asked to determine whether the explanation correctly identifies the underlying event responsible for the changepoint. The judge outputs a binary correctness label (correct/incorrect) based on whether the core causal event is accurately attributed. The exact evaluation prompt is provided in Appendix~\ref{app:llm_prompts}.

\paragraph{Results.} Table~\ref{tab:comparison} compares detection performance. The ensemble method substantially outperforms auto-selection across all metrics.

\begin{table}[h]
\centering
\small
\begin{tabular*}{\columnwidth}{@{\extracolsep{\fill}}lcccccc@{}}
\toprule
\textbf{Method} & \textbf{TP} & \textbf{FP} & \textbf{FN} & \textbf{Prec} & \textbf{Rec} & \textbf{F1} \\
\midrule
Individual & 3 & 1 & 4 & .750 & .429 & .545 \\
Ensemble & 6 & 4 & 1 & .600 & \best{.857} & \best{.706} \\
\bottomrule
\end{tabular*}
\caption{Detection performance: Individual vs.\ Ensemble (7 datasets).}
\label{tab:comparison}
\end{table}

The ensemble detects twice as many true positives (6 vs.\ 3) with higher recall (.857 vs.\ .429) and F1 (.706 vs.\ .545). It also achieves better Mean Temporal Precision (MTE = 0.50 vs.\ 0.67 data points).

\paragraph{LLM Explanation Quality.} For correctly detected breaks, we evaluate whether LLMs identify the ground truth event (Table~\ref{tab:llm_comparison}).

\begin{table}[h]
\centering
\small
\begin{tabular*}{\columnwidth}{@{\extracolsep{\fill}}lcc@{}}
\toprule
\textbf{LLM} & \textbf{Individual} & \textbf{Ensemble} \\
\midrule
GPT-4o & 1/3 (33\%) & 2/6 (33\%) \\
Llama 3.1 8B & 1/3 (33\%) & 4/6 (67\%) \\
DeepSeek-R1 & 1/3 (33\%) & 4/6 (67\%) \\
\midrule
\textbf{Overall} & 3/9 (33\%) & 10/18 (\best{56\%}) \\
\bottomrule
\end{tabular*}
\caption{LLM explanation accuracy by method.}
\label{tab:llm_comparison}
\end{table}

The ensemble method yields higher explanation accuracy (56\% vs.\ 33\%).

\paragraph{End-to-End Performance.} Table~\ref{tab:e2e_comparison} shows the critical metric: correct detection \textit{and} correct explanation.

\begin{table}[h]
\centering
\small
\resizebox{\columnwidth}{!}{%
\begin{tabular}{lcccc}
\toprule
\textbf{Method} & \textbf{GPT-4o} & \textbf{Llama3} & \textbf{DeepSeek} & \textbf{Total} \\
\midrule
Individual & 1/7 & 1/7 & 1/7 & 3/21 (14\%)\\
Ensemble & 2/7 & 4/7 & 4/7 & 10/21 (\best{48\%}) \\
\bottomrule
\end{tabular}%
}
\caption{End-to-end success (correct detection and explanation).}
\label{tab:e2e_comparison}
\end{table}

The ensemble achieves 3.3$\times$ higher end-to-end success rate (48\% vs.\ 14\%), demonstrating that the benefits compound: better detection \textit{and} better event-attribution.

\subsection{RAG}
\label{sec:rag_eval}

To demonstrate our RAG pipeline, we construct a synthetic scenario with controlled ground truth: a fictional company (Nexora Technologies) tracking monthly active users from 2020--2024. The ground truth changepoint occurs in July 2022 when the company launched ``Project Helios''---an AI-powered recommendation engine that increased user engagement by 40\% and caused monthly active users to surge from 175,000 to over 210,000. The corpus contains 31 documents: 30 distractor documents (HR policies, IT notices, unrelated meeting notes) and 1 relevant internal memo announcing the Project Helios launch. Example documents are shown in Appendix~\ref{app:rag_documents}.

\paragraph{Experimental Conditions.} We test two conditions: (1) \textbf{With Relevant Document}: RAG retrieves from the full corpus including the relevant document; (2) \textbf{Clutter Only}: RAG retrieves only from distractor documents.

\paragraph{Expected Behavior.} With the relevant document available, the system should correctly identify ``Project Helios Launch'' as the cause. With only clutter documents, the system should acknowledge uncertainty rather than hallucinate an explanation.

\paragraph{Results.} In condition (1), the system correctly retrieves the Project Helios memo and generates an accurate explanation citing the specific event, date, and key personnel. In condition (2), the system appropriately indicates that the retrieved documents do not contain information explaining the observed changepoint, demonstrating calibrated uncertainty.

\section{Discussion}

The ensemble method's advantage stems from two complementary factors. First, aggregating multiple detection algorithms reduces sensitivity to individual method failures---different algorithms have different strengths, and if one method misses a break due to its assumptions being violated, others may still detect it. Our results show the ensemble detects twice as many true positives as automatic method selection (6 vs.\ 3), precisely because single methods do not excels across all dataset characteristics.

The $\sim$50\% explanation accuracy without RAG reflects the fundamental challenge of attributing statistical changes to specific historical events using only parametric knowledge. "Niche" events like ``FCC Robocall Ruling (March 2018)'' or ``Peak Antarctic Ozone Depletion (1993)'' may not be well-represented in LLM training data. Notably, all three LLMs -- despite varying in size from 8B to frontier-scale -- performed similarly, suggesting this limitation is not simply a matter of model capacity but rather knowledge coverage.

Our RAG extension addresses this gap for private or specialized domains. The synthetic evaluation demonstrates that when relevant documents exist, the system correctly grounds its explanations; when they do not, it appropriately expresses uncertainty rather than hallucinating plausible-sounding but incorrect explanations.

\section{Conclusion}

We present a framework that bridges statistical changepoint detection and natural language explanation through ensemble methods and LLM integration. Our evaluation on seven benchmark datasets demonstrates three key findings:

\begin{enumerate}
    \item \textbf{Ensemble detection outperforms individual methods}, achieving higher recall (0.857 vs.\ 0.429) and F1 score (0.706 vs.\ 0.545) compared to automatic method selection.

    \item \textbf{Richer detection output improves explanations}. The ensemble's confidence scores and method agreement information help LLMs generate more accurate explanations (56\% vs.\ 33\% accuracy).

    \item \textbf{End-to-end evaluation matters}. The ensemble achieves 3.3$\times$ higher end-to-end success rate, demonstrating that detection and explanation quality compound rather than being independent.
\end{enumerate}

The RAG extension enables reliable explanations for private data by grounding LLM outputs in user-provided documents while maintaining calibrated uncertainty when relevant context is unavailable.

\section*{Limitations}

Our framework has several limitations.

\paragraph{Scope.} The current implementation supports only univariate time series and post-hoc explanation. Future work should extend the framework to online changepoint detection for streaming data and multivariate time series analysis for complex, high-dimensional systems.

\paragraph{Evaluation Scale.} Our evaluation uses seven benchmark datasets with single known changepoints. While sufficient for demonstrating the framework's capabilities, this limited sample size precludes strong statistical claims about generalization.

\paragraph{Confidence Scores.} Confidence scores at both levels are heuristic rather than statistically rigorous. At the individual method level, confidence calculations vary: some derive from test statistics scaled by critical values (CUSUM), others from p-value transformations (Bai-Perron: $1 - p_{\text{value}}$), and others from heuristics like local variance reduction (Binary Segmentation). These heterogeneous measures are not calibrated probabilities. At the ensemble level, the voting-based aggregation reflects method agreement rather than true probabilistic uncertainty. Unlike Bayesian approaches such as BEAST \cite{zhao2019beast} that yield principled posterior distributions, our metrics prioritize interpretability and computational efficiency over statistical formalism. See Appendix~\ref{app:cpd_methods} for detailed confidence calculations per method.

\paragraph{LLM Limitations.} Explanation quality depends on the LLM's capabilities and knowledge. The risk of hallucination persists despite prompt engineering and RAG grounding; human oversight remains necessary for critical applications. Additionally, using cloud-based LLM APIs requires transmitting data to external servers, which may be unsuitable for sensitive or proprietary data. Local model deployment or the RAG pipeline with on-premise infrastructure can mitigate this concern.

\paragraph{Computational Cost.} Running multiple detection methods with subsequent clustering is computationally expensive compared to single-method approaches. For very long time series, this overhead may be prohibitive, though users can select method subsets to reduce cost.

\paragraph{Parameter Sensitivity.} Performance depends on parameter choices: clustering tolerance, minimum vote thresholds, and RAG retrieval windows all require tuning based on data characteristics and domain knowledge.

\bibliography{custom}

\appendix

\section{Changepoint Detection Method Details}
\label{app:cpd_methods}

This appendix provides detailed implementation information for all twelve changepoint detection methods used in our ensemble framework, including the underlying packages, algorithms, parameters, and confidence score calculations.

\subsection{Statistical Methods}

These methods are implemented using custom code built on top of \texttt{statsmodels} and \texttt{scipy}.

\subsubsection{CUSUM (Cumulative Sum)}

\textbf{Implementation:} Custom implementation using statsmodels.stats.diagnostic.breaks\_cusumolsresid for the core CUSUM test.

\textbf{Minimum Data:} 15 observations.

\textbf{Parameters:}
\begin{itemize}[nosep]
    \item \texttt{significance\_level}: Significance level for break detection (default: 0.05)
    \item \texttt{trend}: Trend specification---\texttt{'n'} (none), \texttt{'c'} (constant), \texttt{'ct'} (constant + trend); default: \texttt{'c'}
    \item \texttt{use\_recursive}: Whether to use recursive CUSUM for multiple breaks (default: False)
\end{itemize}

\textbf{Algorithm:}
\begin{enumerate}[nosep]
    \item Fit OLS regression based on trend specification: $y_t = X\beta + \epsilon_t$
    \item Calculate residuals: $\hat{\epsilon}_t = y_t - X\hat{\beta}$
    \item Compute CUSUM statistic: $S_t = \sum_{i=1}^{t} (\hat{\epsilon}_i - \bar{\epsilon})$
    \item Calculate scaled statistic: $S_t^* = S_t / (\hat{\sigma}\sqrt{n})$
    \item Find break at $t^* = \arg\max_t |S_t^*|$
    \item Test significance using Brown-Durbin-Evans critical values
\end{enumerate}

\textbf{Confidence:} $c = \min(0.95, \max(0.1, |S_{t^*}^*| / c_{\alpha}))$ where $c_{\alpha} = 1.36$ (5\% significance), $1.63$ (1\%), or $1.14$ (10\%).

\subsubsection{Bai-Perron Test}

\textbf{Implementation:} Custom implementation based on the Bai-Perron sequential testing procedure with dynamic programming for optimal partition.

\textbf{Minimum Data:} 10 observations.

\textbf{Parameters:}
\begin{itemize}[nosep]
    \item \texttt{max\_breaks}: Maximum number of breaks to detect (default: 5)
    \item \texttt{min\_segment\_size}: Minimum observations per segment (default: 15\% of data)
    \item \texttt{significance\_level}: Significance level for F-test (default: 0.05)
    \item \texttt{h}: Trimming parameter (default: 0.15)
\end{itemize}

\textbf{Algorithm:}
\begin{enumerate}[nosep]
    \item Filter methods based on minimum segment size requirements
    \item For $m = 1, \ldots, M$ breaks, minimize global SSR using dynamic programming
    \item Calculate F-statistic: $F = \frac{(\text{SSR}_{\text{restricted}} - \text{SSR}_{\text{unrestricted}}) / k}{\text{SSR}_{\text{unrestricted}} / (n - 2k - 1)}$
    \item Test significance using F-distribution critical values
    \item Stop when additional breaks are no longer significant
\end{enumerate}

\textbf{Confidence:} $c = 1 - p_{\text{value}}$ where $p_{\text{value}}$ is from the F-distribution CDF.

\subsubsection{Chow Test}

\textbf{Implementation:} Custom implementation using \texttt{statsmodels.api.OLS} for regression fitting.

\textbf{Minimum Data:} 20 observations.

\textbf{Parameters:}
\begin{itemize}[nosep]
    \item \texttt{significance\_level}: Significance level (default: 0.05)
    \item \texttt{trend}: Trend specification (default: \texttt{'ct'})
    \item \texttt{search\_method}: \texttt{'grid'} (exhaustive) or \texttt{'sequential'} (recursive); default: \texttt{'grid'}
    \item \texttt{min\_segment\_size}: Minimum segment size (default: 15\% of data)
    \item \texttt{test\_multiple\_points}: Whether to detect multiple breaks (default: True)
\end{itemize}

\textbf{Algorithm:}
\begin{enumerate}[nosep]
    \item For each candidate break point $\tau$, split data into $[1, \tau]$ and $[\tau+1, n]$
    \item Fit three OLS regressions: full sample, first segment, second segment
    \item Calculate Chow F-statistic: $F = \frac{(\text{RSS}_{\text{full}} - \text{RSS}_1 - \text{RSS}_2) / k}{(\text{RSS}_1 + \text{RSS}_2) / (n - 2k)}$
    \item Test against $F_{k, n-2k}$ distribution
    \item Select breaks with $p < \alpha$, ensuring minimum separation
\end{enumerate}

\textbf{Confidence:} $c = \max(0.05, \min(0.95, 1 - p_{\text{value}}))$.

\subsubsection{Zivot-Andrews Test}

\textbf{Implementation:} Uses the \texttt{arch} package (\texttt{arch.unitroot.ZivotAndrews}).

\textbf{Minimum Data:} 20 observations.

\textbf{Parameters:}
\begin{itemize}[nosep]
    \item \texttt{trend}: Break type---\texttt{'c'} (intercept break), \texttt{'t'} (trend break), \texttt{'ct'} (both); default: \texttt{'c'}
    \item \texttt{lags}: Number of lags to include (default: automatic selection)
\end{itemize}

\textbf{Algorithm:}
\begin{enumerate}[nosep]
    \item Run Zivot-Andrews unit root test with structural break
    \item If test rejects unit root ($p < 0.05$), search for optimal break location
    \item For each candidate break, fit augmented Dickey-Fuller regression with break dummy
    \item Select break that minimizes the unit root test statistic (most negative $t$-statistic)
\end{enumerate}

\textbf{Confidence:} $c = \max(0, 1 - p_{\text{value}})$ from the Zivot-Andrews test.

\subsection{Changepoint Methods (Ruptures-based)}

These methods use the \texttt{ruptures} package \cite{truong2020ruptures} with custom preprocessing and confidence calculations.

\subsubsection{PELT (Pruned Exact Linear Time)}

\textbf{Implementation:} Uses \texttt{ruptures.Pelt} with model-aware preprocessing.

\textbf{Minimum Data:} 10 observations.

\textbf{Parameters:}
\begin{itemize}[nosep]
    \item \texttt{model}: Cost function---\texttt{'l2'} (squared error), \texttt{'l1'} (absolute), \texttt{'rbf'} (kernel), \texttt{'linear'}, \texttt{'normal'}, \texttt{'ar'}; default: \texttt{'l2'}
    \item \texttt{penalty}: Penalty value $\beta$ controlling number of changepoints (default: $3\log n$)
    \item \texttt{n\_bkps}: Fixed number of breakpoints (overrides penalty if set)
    \item \texttt{min\_size}: Minimum segment length (default: $\max(2, 0.02n)$)
    \item \texttt{jump}: Subsampling interval (default: 1)
\end{itemize}

\textbf{Algorithm:} Minimize penalized cost $\sum_{i=0}^{m} c(y_{t_i:t_{i+1}}) + \beta m$ using dynamic programming with pruning, achieving $O(n)$ complexity under certain conditions.

\textbf{Confidence:} $c = 1 - \exp(-z)$ where $z = |\bar{y}_{\text{right}} - \bar{y}_{\text{left}}| / \sigma_{\text{local}}$ using a 5-point window around each break.

\subsubsection{Binary Segmentation}

\textbf{Implementation:} Uses \texttt{ruptures.Binseg}.

\textbf{Minimum Data:} 10 observations.

\textbf{Parameters:}
\begin{itemize}[nosep]
    \item \texttt{model}: Cost function (default: \texttt{'l2'})
    \item \texttt{n\_bkps}: Fixed number of breakpoints (optional)
    \item \texttt{penalty}: Penalty value (default: $2\log n$ BIC penalty)
    \item \texttt{jump}: Subsampling interval (default: 5)
    \item \texttt{min\_size}: Minimum segment length (default: 2)
\end{itemize}

\textbf{Algorithm:} Recursively find single changepoint that maximizes cost reduction, then apply to each resulting segment until stopping criterion is met.

\textbf{Confidence:} Based on local variance reduction: $c = \min(0.95, \max(0.1, 2 \cdot (\sigma^2_{\text{total}} - \sigma^2_{\text{segmented}}) / \sigma^2_{\text{total}}))$.

\subsubsection{Dynamic Programming (Optimal Partitioning)}

\textbf{Implementation:} Uses \texttt{ruptures.Dynp}.

\textbf{Minimum Data:} 10 observations (warning issued for $n > 2000$ due to $O(Qn^2)$ complexity).

\textbf{Parameters:}
\begin{itemize}[nosep]
    \item \texttt{model}: Cost function (default: \texttt{'l2'})
    \item \texttt{n\_bkps}: Number of breakpoints (required; estimated from penalty if not provided)
    \item \texttt{jump}: Subsampling interval (default: 1)
    \item \texttt{min\_size}: Minimum segment length (default: 2)
\end{itemize}

\textbf{Algorithm:} Find globally optimal segmentation by exhaustive dynamic programming over all possible partitions.

\textbf{Confidence:} $c = \min(0.95, \max(0.15, 0.3 + 0.6 \cdot \text{local\_improvement}))$ where local improvement is the cost reduction ratio.

\subsubsection{MOSUM (Moving Sum)}

\textbf{Implementation:} Custom MOSUM algorithm (not using ruptures internally).

\textbf{Minimum Data:} 20 observations.

\textbf{Parameters:}
\begin{itemize}[nosep]
    \item \texttt{window}: Window size for moving sum (default: $\max(10, n/10)$)
    \item \texttt{n\_bkps}: Fixed number of breakpoints (optional)
    \item \texttt{penalty}: Penalty for automatic selection (optional)
\end{itemize}

\textbf{Algorithm:}
\begin{enumerate}[nosep]
    \item For each position $k \in [\text{window}, n - \text{window}]$:
    \item Compute left/right window means and pooled variance
    \item Calculate MOSUM statistic: $T_k = |\bar{y}_{\text{right}} - \bar{y}_{\text{left}}| \sqrt{w / (2\hat{\sigma}^2_{\text{pooled}})}$
    \item Select breaks where $T_k > 3.5$ (approximately 5\% critical value)
    \item Merge breaks within window$/2$ distance
\end{enumerate}

\textbf{Confidence:} $c = 0.5 + 0.4 \cdot \min(1, w/20) + 0.1 \cdot \min(1, d_{\text{boundary}}/w)$.

\subsubsection{Wild Binary Segmentation (WBS)}

\textbf{Implementation:} Custom WBS using \texttt{ruptures.Binseg} on random intervals.

\textbf{Minimum Data:} 30 observations.

\textbf{Parameters:}
\begin{itemize}[nosep]
    \item \texttt{width}: Relative width of random intervals (default: 0.05 = 5\% of data)
    \item \texttt{n\_bkps}: Fixed number of breakpoints (optional)
    \item \texttt{penalty}: Penalty for automatic selection (optional)
\end{itemize}

\textbf{Algorithm:}
\begin{enumerate}[nosep]
    \item Generate $\max(100, 2n)$ random intervals of width $[w, 2w]$ where $w = \max(10, 0.05n)$
    \item Apply binary segmentation to each interval, recording detected break locations
    \item Aggregate break candidates by frequency of detection across intervals
    \item Select breaks detected in $\geq 5\%$ of intervals (or top $k$ if \texttt{n\_bkps} specified)
\end{enumerate}

\textbf{Confidence:} $c = 0.65 + 0.25 \cdot \min(1, d_{\text{boundary}} / w)$ where $d_{\text{boundary}}$ is distance to nearest boundary.

\subsection{Machine Learning Methods}

\subsubsection{Prophet}

\textbf{Implementation:} Uses Meta's \texttt{prophet} package.

\textbf{Minimum Data:} 30 observations.

\textbf{Parameters:}
\begin{itemize}[nosep]
    \item \texttt{n\_changepoints}: Number of potential changepoint locations (default: 25)
    \item \texttt{changepoint\_range}: Proportion of history for changepoint inference (default: 0.8)
    \item \texttt{changepoint\_prior\_scale}: Regularization strength---smaller values yield fewer changepoints (default: 0.02)
\end{itemize}

\textbf{Algorithm:} Prophet fits an additive model $y(t) = g(t) + s(t) + h(t) + \epsilon_t$ where $g(t)$ is a piecewise linear trend with changepoints placed using a sparse Laplace prior. Changepoints with $|\delta_j| > 0.01 \cdot \text{std}(y)$ are considered significant.

\textbf{Confidence:} $c = 0.4 + 0.5 \cdot (|\delta_j| / \max_i |\delta_i|)$ based on relative changepoint magnitude.

\section{Automatic Method Selection: Complete Specification}
\label{app:auto_selection_details}

This appendix provides the complete specification of our automatic method selection algorithm, including exact data requirements and scoring matrices.

\subsection{Method-Specific Data Requirements}

\begin{table}[H]
\centering
\resizebox{\columnwidth}{!}{%
\begin{tabular}{lcc}
\toprule
\textbf{Method} & \textbf{Min. Points} & \textbf{Rationale} \\
\midrule
Bai-Perron & 10 & Requires sufficient segments \\
CUSUM & 15 & Needs baseline for cumsum \\
Chow Test & 20 & Requires two segments \\
Zivot-Andrews & 20 & ADF test with break term \\
PELT & 10 & DP initialization \\
Binary Seg. & 10 & Recursive splitting \\
Dynamic Prog. & 10 & Optimal partitioning \\
MOSUM & 20 & Moving window \\
Wild Binary Seg. & 30 & Random interval generation \\
Prophet & 30 & Trend + seasonality fitting \\
\bottomrule
\end{tabular}%
}
\caption{Minimum data size requirements for each changepoint detection method.}
\label{tab:method_requirements}
\end{table}

\subsection{Scoring Criterion 1: Sample Size Suitability}

The sample size suitability function $f_1(m, n)$ assigns scores based on each method's optimal operating range.

\begin{table}[H]
\centering
\resizebox{\columnwidth}{!}{%
\begin{tabular}{lccc}
\toprule
\textbf{Method} & \textbf{Small ($n < 50$)} & \textbf{Medium ($50 \leq n < 1000$)} & \textbf{Large ($n \geq 1000$)} \\
\midrule
Bai-Perron & 0.3 & 0.9 & 0.6 \\
CUSUM & 0.9 (if $n \geq 20$), else 0.2 & 0.9 & 0.9 \\
Chow Test & 0.8 (if $n \geq 40$), else 0.4 & 0.8 & 0.8 \\
Zivot-Andrews & 0.8 (if $n \geq 30$), else 0.3 & 0.8 & 0.8 \\
PELT & 0.6 & 0.9 & 0.9 \\
Binary Seg. & 0.8 (if $n \geq 30$), else 0.5 & 0.8 & 0.8 \\
Dynamic Prog. & 0.4 & 0.7 & 0.7 \\
MOSUM & 0.8 (if $n \geq 40$), else 0.3 & 0.8 & 0.8 \\
Wild Binary Seg. & 0.4 & 0.8 (if $n \geq 100$), else 0.4 & 0.8 \\
Prophet & 0.4 (if $n \geq 50$), else 0.1 & 0.9 (if $n \geq 100$), else 0.4 & 0.9 \\
\bottomrule
\end{tabular}%
}
\caption{Sample size suitability scores $f_1(m, n)$.}
\label{tab:sample_size_scores}
\end{table}

\subsection{Scoring Criterion 2: Noise Tolerance}

The noise tolerance function $f_2(m, \nu)$ evaluates robustness to noise, where $\nu = \sigma/|\mu|$ is the coefficient of variation.

\begin{table}[H]
\centering
\resizebox{\columnwidth}{!}{%
\begin{tabular}{lccc}
\toprule
\textbf{Method} & \textbf{Clean ($\nu < 0.2$)} & \textbf{Moderate ($0.2 \leq \nu < 0.5$)} & \textbf{High ($\nu \geq 0.5$)} \\
\midrule
Bai-Perron & 0.9 & 0.6 & 0.3 \\
CUSUM & 0.7 & 0.8 & 0.6 \\
Chow Test & 0.8 & 0.7 & 0.4 \\
Zivot-Andrews & 0.8 & 0.6 & 0.4 \\
PELT & 0.8 & 0.9 & 0.7 \\
Binary Seg. & 0.7 & 0.8 & 0.7 \\
Dynamic Prog. & 0.8 & 0.8 & 0.6 \\
MOSUM & 0.6 & 0.7 & 0.6 \\
Wild Binary Seg. & 0.5 & 0.8 & 0.9 \\
Prophet & 0.6 & 0.8 & 0.8 \\
\bottomrule
\end{tabular}%
}
\caption{Noise tolerance scores $f_2(m, \nu)$.}
\label{tab:noise_scores}
\end{table}

\subsection{Scoring Criterion 3: Trend Handling}

Trend strength $\rho$ is measured by Pearson correlation between time index and values.

\begin{table}[H]
\centering
\resizebox{\columnwidth}{!}{%
\begin{tabular}{lccc}
\toprule
\textbf{Method} & \textbf{No Trend ($\rho < 0.2$)} & \textbf{Moderate ($0.2 \leq \rho < 0.6$)} & \textbf{Strong ($\rho \geq 0.6$)} \\
\midrule
Bai-Perron & 0.7 & 0.7 & 0.5 \\
CUSUM & 0.7 & 0.8 & 0.6 \\
Chow Test & 0.7 & 0.8 & 0.6 \\
Zivot-Andrews & 0.7 & 0.6 & 0.4 \\
PELT & 0.7 & 0.7 & 0.5 \\
Binary Seg. & 0.7 & 0.7 & 0.5 \\
Dynamic Prog. & 0.7 & 0.7 & 0.5 \\
MOSUM & 0.7 & 0.7 & 0.6 \\
Wild Binary Seg. & 0.7 & 0.6 & 0.4 \\
Prophet & 0.7 & 0.9 & 1.0 \\
\bottomrule
\end{tabular}%
}
\caption{Trend handling scores $f_3(m, \rho)$.}
\label{tab:trend_scores}
\end{table}

\subsection{Scoring Criterion 4: Seasonality Handling}

Seasonality strength $\lambda$ is estimated from autocorrelation at seasonal lags.

\begin{table}[H]
\centering
\resizebox{\columnwidth}{!}{%
\begin{tabular}{lcc}
\toprule
\textbf{Method} & \textbf{Low/No ($\lambda < 0.5$)} & \textbf{Strong ($\lambda \geq 0.5$)} \\
\midrule
Bai-Perron & 0.7 & 0.4 \\
CUSUM & 0.7 & 0.5 \\
Chow Test & 0.7 & 0.5 \\
Zivot-Andrews & 0.7 & 0.3 \\
PELT & 0.7 & 0.6 \\
Binary Seg. & 0.7 & 0.6 \\
Dynamic Prog. & 0.7 & 0.6 \\
MOSUM & 0.7 & 0.5 \\
Wild Binary Seg. & 0.7 & 0.5 \\
Prophet & 0.7 & 0.9 \\
\bottomrule
\end{tabular}%
}
\caption{Seasonality handling scores $f_4(m, \lambda)$.}
\label{tab:seasonality_scores}
\end{table}

\subsection{Scoring Criterion 5: Computational Efficiency}

\begin{table}[H]
\centering
\resizebox{\columnwidth}{!}{%
\begin{tabular}{lccc}
\toprule
\textbf{Method} & \textbf{Small ($n < 100$)} & \textbf{Medium ($100 \leq n < 1000$)} & \textbf{Large ($n \geq 1000$)} \\
\midrule
Bai-Perron & 0.7 & 0.6 & 0.4 \\
CUSUM & 0.7 & 0.9 & 0.8 \\
Chow Test & 0.7 & 0.7 & 0.5 \\
Zivot-Andrews & 0.7 & 0.8 & 0.6 \\
PELT & 0.7 & 0.9 & 1.0 \\
Binary Seg. & 0.7 & 0.8 & 0.9 \\
Dynamic Prog. & 0.7 & 0.6 & 0.4 \\
MOSUM & 0.7 & 0.7 & 0.6 \\
Wild Binary Seg. & 0.7 & 0.5 & 0.3 \\
Prophet & 0.7 & 0.7 & 0.6 \\
\bottomrule
\end{tabular}%
}
\caption{Computational efficiency scores $f_5(m, n)$.}
\label{tab:efficiency_scores}
\end{table}

\subsection{Scoring Criterion 6: Stationarity Handling}

Stationarity is determined by ADF test $p$-value $s$.

\begin{table}[H]
\centering
\resizebox{\columnwidth}{!}{%
\begin{tabular}{lcc}
\toprule
\textbf{Method} & \textbf{Stationary ($s \leq 0.05$)} & \textbf{Non-Stationary ($s > 0.05$)} \\
\midrule
Bai-Perron & 0.9 & 0.3 \\
CUSUM & 0.8 & 0.5 \\
Chow Test & 0.8 & 0.4 \\
Zivot-Andrews & 0.6 & 1.0 \\
PELT & 0.8 & 0.6 \\
Binary Seg. & 0.8 & 0.6 \\
Dynamic Prog. & 0.8 & 0.6 \\
MOSUM & 0.8 & 0.5 \\
Wild Binary Seg. & 0.7 & 0.5 \\
Prophet & 0.7 & 0.8 \\
\bottomrule
\end{tabular}%
}
\caption{Stationarity handling scores $f_6(m, s)$.}
\label{tab:stationarity_scores}
\end{table}

\subsection{Scoring Criterion 7: Outlier Robustness}

Outlier ratio $o$ is computed using the IQR criterion.

\begin{table}[H]
\centering
\resizebox{\columnwidth}{!}{%
\begin{tabular}{lcc}
\toprule
\textbf{Method} & \textbf{Few Outliers ($o < 0.05$)} & \textbf{Many Outliers ($o \geq 0.05$)} \\
\midrule
Bai-Perron & 0.7 & 0.3 \\
CUSUM & 0.7 & 0.6 \\
Chow Test & 0.7 & 0.4 \\
Zivot-Andrews & 0.7 & 0.4 \\
PELT & 0.7 & 0.7 \\
Binary Seg. & 0.7 & 0.7 \\
Dynamic Prog. & 0.7 & 0.7 \\
MOSUM & 0.7 & 0.6 \\
Wild Binary Seg. & 0.7 & 0.9 \\
Prophet & 0.7 & 0.8 \\
\bottomrule
\end{tabular}%
}
\caption{Outlier robustness scores $f_7(m, o)$.}
\label{tab:outlier_scores}
\end{table}

\subsection{Scoring Rationale}

The scoring functions are heuristic mappings derived from the statistical literature on each method's theoretical properties and empirical performance. These scores encode approximate expert knowledge rather than theoretically optimal weightings:

\begin{itemize}[nosep]
    \item \textbf{Sample Size:} Methods requiring complex model fitting (Bai-Perron, Prophet) need larger samples. PELT and CUSUM are more flexible across sizes.
    \item \textbf{Noise Tolerance:} Robust methods (WBS, PELT, Prophet) score higher under noisy conditions due to regularization or resampling. Parametric tests require cleaner data.
    \item \textbf{Trend Handling:} Prophet's additive trend model excels; traditional methods assume detrended data.
    \item \textbf{Seasonality:} Prophet explicitly models seasonality via Fourier terms; other methods may conflate seasonal fluctuations with changepoints.
    \item \textbf{Computational Efficiency:} PELT has $O(n)$ complexity. Dynamic Programming and Bai-Perron are expensive for large $n$.
    \item \textbf{Stationarity:} Zivot-Andrews is designed for unit root breaks in non-stationary data.
    \item \textbf{Outlier Robustness:} WBS's random interval approach and Prophet's outlier detection make them robust.
\end{itemize}

\section{LLM Prompt Templates}
\label{app:llm_prompts}

This appendix provides the exact prompt templates used for LLM-based changepoint explanation in our framework.

\subsection{Standard Explanation Mode}

This mode is used when no domain-specific documents are provided.

\begin{systempromptbox}[System Prompt]
You are a data analyst expert in time series analysis.

Your task is to explain structural breaks - significant, persistent changes in time series data.

Provide clear, concise explanations that:
1. Describe what changed (magnitude and direction)
2. Suggest possible causes based on the timing and statistical evidence
3. Think of possible external events near the break date (e.g. macro, policy, company news..), flagging speculation if unsure
4. Assess the significance of the change
5. Avoid speculation beyond what the data supports

Be specific and professional.
\end{systempromptbox}

\begin{userpromptbox}[User Prompt Template]
Analyze this structural break in \{data\_description\}:

Break Details:
- Date: \{break\_date\}
- Confidence: \{confidence:.1\%\}
- Magnitude: \{magnitude:.2f\} (\{direction\} shift)

Before Break (30-day window):
- Mean: \{before\_stats['mean']:.2f\}
- Std Dev: \{before\_stats['std']:.2f\}
- Trend: \{before\_stats['trend']\}

After Break (30-day window):
- Mean: \{after\_stats['mean']:.2f\}
- Std Dev: \{after\_stats['std']:.2f\}
- Trend: \{after\_stats['trend']\}

Provide a brief explanation of this structural break.
\end{userpromptbox}

\textbf{Generation Parameters:} Temperature: 0.3 (low for factual, consistent explanations); Max Tokens: 300; Window Size: 30 days before and after break (configurable).

\subsection{RAG-Enhanced Explanation Mode}

This mode is used when domain-specific documents are provided.

\begin{systempromptbox}[System Prompt (RAG Mode)]
You are a data analyst expert in time series analysis.

You have access to relevant documents that may explain the structural break.

When explaining:
1. Connect the statistical evidence to events in the documents
2. Be specific about which information supports your explanation
3. Distinguish between correlation and likely causation
4. Keep explanations concise and actionable
\end{systempromptbox}

\begin{userpromptbox}[User Prompt Template (RAG Mode)]
Analyze this structural break with additional context:

Break Information:
- Date: \{break\_date\}
- Confidence: \{confidence\}
- Magnitude: \{magnitude\}
- Direction: \{direction\}

Relevant Documents:
\{document\_context\}

Explain this break using both the statistical evidence and document context. Be specific about how the documents relate to the observed change.
\end{userpromptbox}

\textbf{RAG Parameters:} Temperature: 0.3; Max Tokens: 400; Top-k: 3 most relevant document chunks; Embedding Model: Sentence-Transformers (all-MiniLM-L6-v2); Vector Store: ChromaDB with cosine similarity.

\subsection{Judge Evaluation Prompt}

This prompt is used to evaluate whether an LLM-generated explanation correctly identifies the ground truth event responsible for the changepoint.

\begin{systempromptbox}[Judge System Prompt]
You are an expert evaluator assessing the quality of changepoint explanations. Your task is to determine whether a generated explanation correctly identifies the underlying event that caused a structural break in time series data.

You will receive:
1. The LLM's explanation of a detected changepoint
2. The ground truth event that actually caused the changepoint

Evaluate whether the explanation correctly identifies the core causal event. The explanation does not need to match the ground truth word-for-word, but must identify the same fundamental event or cause.

Output only: CORRECT or INCORRECT
\end{systempromptbox}

\begin{userpromptbox}[Judge User Prompt Template]
Evaluate the following changepoint explanation:

\textbf{LLM Explanation:}
\{llm\_explanation\}

\textbf{Ground Truth Event:}
\{ground\_truth\_event\}

Does the explanation correctly identify the event that caused the changepoint?
Output only: CORRECT or INCORRECT
\end{userpromptbox}

\textbf{Judge Parameters:} Model: Kimi K2 (1T parameters, 32B active); Temperature: 0.0 (deterministic evaluation).

\section{Usage Workflow Examples}
\label{app:workflow_examples}

This appendix demonstrates practical usage through code examples covering the complete analysis pipeline.

\subsection{Time Series Analysis}

\begin{pythonbox}[Time Series Feature Extraction and LLM Explanation]
import pandas as pd
from structural_break_analyzer import TimeSeriesAnalyzer, LLMExplainer

# Load and analyze time series
data = pd.read_csv('financial_data.csv')
ts_analyzer = TimeSeriesAnalyzer()
result = ts_analyzer.analyze(data, value_col='value', date_col='date')

# View extracted features
print(result.summary())
result.plot()  # 12-panel diagnostic visualization

# Optional: LLM explanation with detail levels
explainer = LLMExplainer(provider="azure", ...)
explained = explainer.explain_timeseries(
    result,
    data_description="daily stock returns",
    detail_level="medium"  # basic, medium, or detailed
)
print(explained.summary())
\end{pythonbox}

\subsection{Individual Method Detection}

\begin{pythonbox}[Using Specific Detection Methods]
from structural_break_analyzer import StructuralBreakAnalyzer

analyzer = StructuralBreakAnalyzer()

# Use any of the 10 available methods
result = analyzer.detect(data, value_col='value', date_col='date',
    method='pelt'  # Options: pelt, bai_perron, cusum, chow_test,
                   # zivot_andrews, binary_segmentation,
                   # dynamic_programming, mosum,
                   # wild_binary_segmentation, prophet
)
print(result.summary())
\end{pythonbox}

\subsection{Automatic Method Selection}

\begin{pythonbox}[Data-Driven Method Selection]
# Auto-select best method based on data characteristics
result = analyzer.detect(data, value_col='value', date_col='date',
    method='auto'  # Analyzes data and selects optimal method
)

# The selected method is recorded in metadata
print(f"Selected method: {result.metadata['selected_method']}")
print(f"Selection scores: {result.metadata['method_scores']}")
\end{pythonbox}

\subsection{Ensemble Detection}

\begin{pythonbox}[Ensemble-based Changepoint Detection]
# Detect breaks using ensemble method (runs all 10 methods)
result = analyzer.detect(
    data,
    value_col='value',
    date_col='date',
    method='ensemble'  # Runs all 10 methods
)

# Print detected breaks with voting statistics
print(result.summary())
result.plot()  # Visualize breaks
\end{pythonbox}

\begin{codeboxenv}[Ensemble Detection Output]
Structural Break Detection Results
Breaks detected: 3

Break 1: 2020-03-09
  Confidence: 83.3
  Magnitude: 1.893

Break 2: 2020-04-15
  Confidence: 54.8
  Magnitude: 0.705

Break 3: 2020-05-20
  Confidence: 72.9
  Magnitude: -1.198
\end{codeboxenv}

\subsection{LLM-Enhanced Explanations}

\begin{pythonbox}[Generating LLM Explanations]
from structural_break_analyzer import LLMExplainer
import os

# Initialize LLM explainer
explainer = LLMExplainer(
    provider="azure",
    azure_endpoint=os.getenv('AZURE_OPENAI_ENDPOINT'),
    api_key=os.getenv('AZURE_OPENAI_API_KEY'),
    deployment_name="gpt-4"
)

# Generate explanations
explained_result = explainer.explain_breaks(
    result,
    data_description="S&P 500 daily returns",
    detail_level="detailed"
)

# Access explanations
for break_point in explained_result.breaks:
    print(f"Break at {break_point.date}:")
    print(f"  {break_point.explanation}")
\end{pythonbox}

\begin{assistantbox}[LLM-Generated Explanation]
The structural break on March 9, 2020, aligns with the early stages of the COVID-19 pandemic's impact on global financial markets. Markets experienced severe volatility as the WHO declared COVID-19 a pandemic on March 11. The upward shift (magnitude: 1.893) reflects increased market turbulence. Major indices entered bear market territory, and circuit breakers were triggered multiple times during this period.
\end{assistantbox}

\subsection{RAG-Enhanced Explanations}

\begin{pythonbox}[RAG-Enhanced Explanations with Private Documents]
# Enable RAG system with document retrieval
explainer.enable_rag(
    cache_dir=".rag_cache",
    embedding_model='all-MiniLM-L6-v2'
)

# Add private context documents
explainer.add_documents([
    "company_reports/",
    "internal_memos/",
    "market_analysis/"
])

# Generate RAG-enhanced explanations
rag_explained = explainer.explain_breaks(
    result,
    data_description="financial market index",
    use_rag=True  # Enable document retrieval
)
\end{pythonbox}

\begin{assistantbox}[RAG-Enhanced Explanation (3 docs retrieved)]
The upward structural break on March 30, 2020, can be causally linked to the Federal Reserve's emergency measures. According to the retrieved documents, the Fed announced a \$700 billion quantitative easing program on March 15, with interest rates cut to 0.00-0.25\%. The timing aligns with when these actions began influencing market behavior, leading to increased stability and a rise in mean value.
\end{assistantbox}

\subsection{Workflow Summary}

The complete workflow demonstrates the framework's end-to-end capabilities: (1) \textbf{Detection} using ensemble method with voting-based consensus; (2) \textbf{Explanation} via LLM with statistical evidence and temporal grounding; (3) \textbf{RAG Enhancement} with private documents for domain-specific context. This unified pipeline bridges statistical detection and actionable insight.

\section{Additional Framework Capabilities}
\label{app:additional_capabilities}

Beyond the core changepoint detection and explanation pipeline presented in this paper, our framework provides several additional capabilities for comprehensive time series analysis. This appendix demonstrates these features through practical examples.

\subsection{Time Series Diagnostics}

The \texttt{TimeSeriesAnalyzer} class provides general-purpose time series analysis including stationarity tests (ADF, KPSS), trend detection, seasonality analysis, and statistical summaries. These diagnostics can also be explained by the LLM using the \texttt{explain\_timeseries()} method with configurable detail levels (basic, medium, detailed).

\begin{pythonbox}[Time Series Feature Extraction]
from structural_break_analyzer import TimeSeriesAnalyzer

# Initialize analyzer
ts_analyzer = TimeSeriesAnalyzer()

# Analyze time series features
result = ts_analyzer.analyze(
    df, value_col='price', date_col='date'
)

# Access extracted features
print(f"Mean: {result.features.mean:.2f}")
print(f"Trend strength: {result.features.trend_strength:.2f}")
print(f"Seasonality: {result.features.seasonality_strength:.2f}")
print(f"ADF p-value: {result.features.adf_pvalue:.4f}")
\end{pythonbox}

\subsection{Automated Report Generation}

The \texttt{ReportGenerator} class produces comprehensive analysis reports in multiple formats (HTML, PDF, Markdown). Reports combine time series diagnostics, detected changepoints with confidence scores, LLM-generated explanations, and visualizations into a single document suitable for stakeholder communication.

\begin{pythonbox}[Generating Analysis Reports]
from structural_break_analyzer import ReportGenerator

# Initialize report generator
generator = ReportGenerator()

# Generate reports in multiple formats
files = generator.generate_report(
    timeseries_result=ts_result,
    break_result=break_result,
    format=['pdf', 'html'],
    title="Q3 2024 Market Analysis",
    output_path="./reports/"
)

print(f"Generated: {files}")
# {'pdf': './reports/report.pdf', 'html': './reports/report.html'}
\end{pythonbox}

\subsection{Interactive Web Interface}

For users who prefer a graphical interface, the framework includes an \texttt{AnalysisWidget} class that provides an interactive web UI within Jupyter notebooks. The widget supports data upload, method configuration, LLM provider settings, RAG document management, and report generation---all without writing code.

\begin{pythonbox}[Launching the Interactive Widget]
from structural_break_analyzer import AnalysisWidget

# Create and display the widget
widget = AnalysisWidget()
widget.show()  # Opens interactive interface in notebook

# Or serve as a standalone web application
widget.servable()  # For use with Panel serve
\end{pythonbox}

\begin{figure}[H]
\centering
\includegraphics[width=\columnwidth]{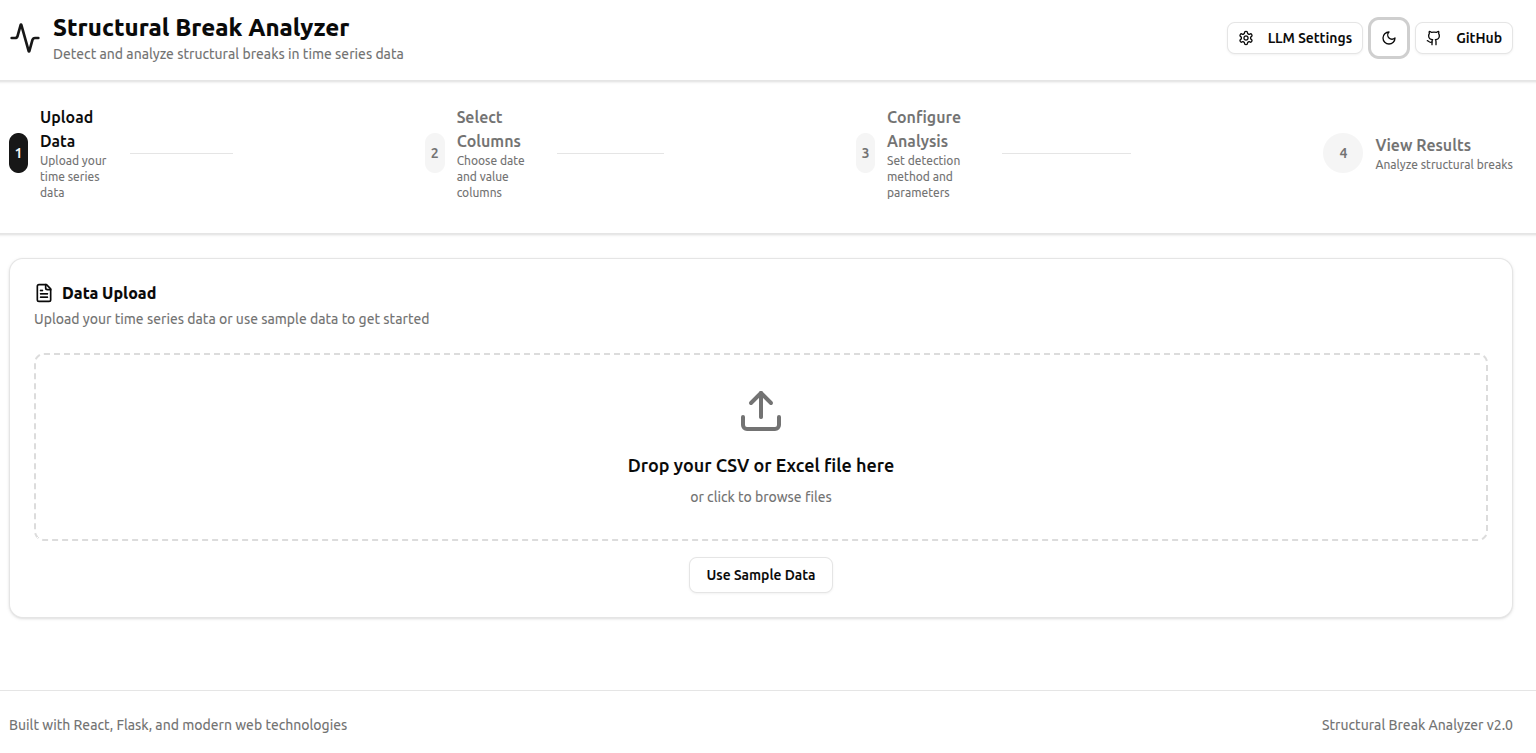}
\caption{Interactive web interface: Step-by-step workflow with data upload via drag-and-drop or sample data, followed by column selection, analysis configuration, and results viewing.}
\label{fig:webui1}
\end{figure}

\begin{figure}[H]
\centering
\includegraphics[width=\columnwidth]{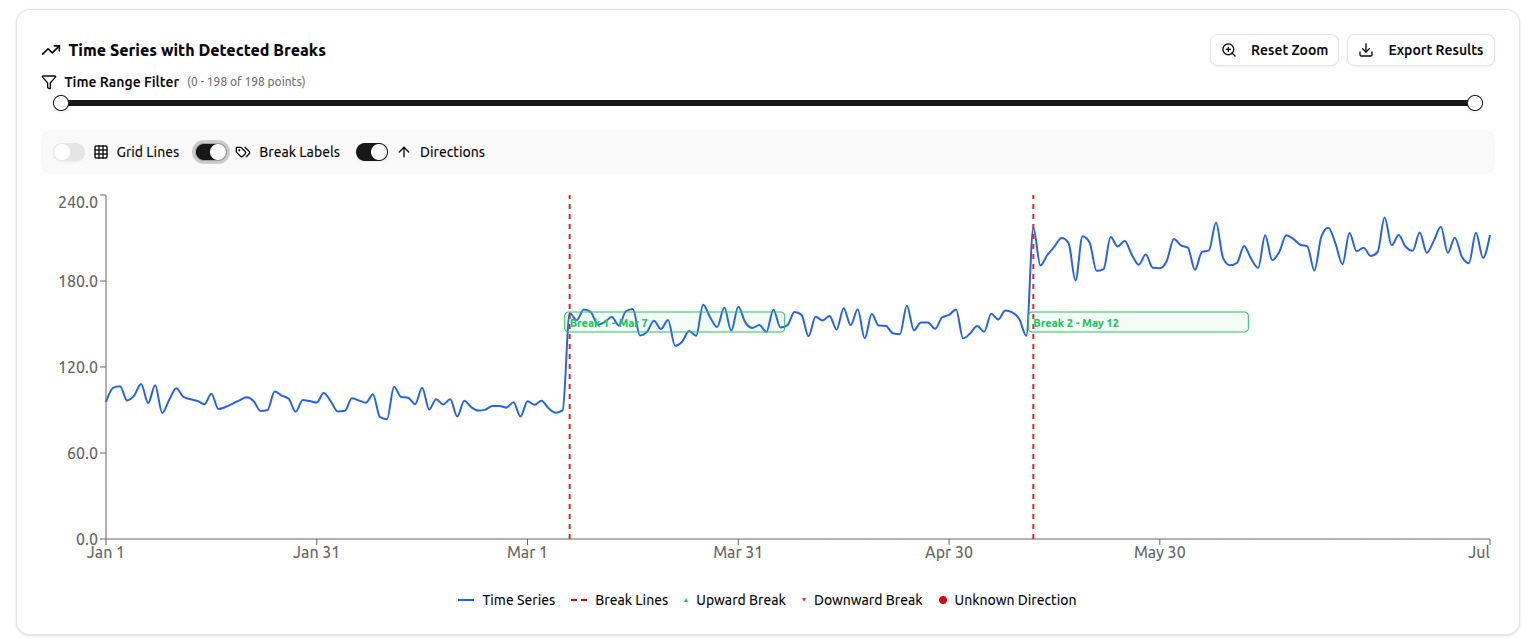}
\caption{Interactive web interface: Results visualization showing detected structural breaks with labeled change directions, time range filtering, and export functionality.}
\label{fig:webui2}
\end{figure}

\subsection{RAG System Management}

The RAG pipeline includes comprehensive management functions: \texttt{get\_rag\_stats()} for monitoring, \texttt{delete\_rag\_documents\_by\_date()} for temporal document filtering, \texttt{clear\_rag\_embedding\_cache()} for cache management, and \texttt{cleanup\_all\_rag\_files()} for complete reset. These enable fine-grained control over the knowledge base.

\begin{pythonbox}[RAG System Management]
# Monitor RAG system status
stats = explainer.get_rag_stats()
print(f"Documents: {stats['total_documents']}")
print(f"Chunks: {stats['total_chunks']}")

# Remove outdated documents by date range
from datetime import datetime
explainer.delete_rag_documents_by_date(
    start_date=datetime(2022, 1, 1),
    end_date=datetime(2022, 6, 30)
)

# Clear embedding cache (keeps documents)
explainer.clear_rag_embedding_cache()

# Complete reset of RAG system
explainer.cleanup_all_rag_files()
\end{pythonbox}

\subsection{Visualization}

All result objects include built-in \texttt{plot()} methods that generate publication-ready visualizations with detected changepoints, confidence intervals, and statistical annotations.

\begin{pythonbox}[Built-in Visualization]
# Plot time series diagnostics (12-panel layout)
fig = ts_result.plot(figsize=(15, 12))
fig.savefig("diagnostics.png", dpi=300)

# Plot break detection results
fig = break_result.plot(figsize=(12, 6))
fig.savefig("breaks.png", dpi=300)

# Access summary for logging
print(break_result.summary())
\end{pythonbox}

\section{RAG Evaluation Documents}
\label{app:rag_documents}

This appendix shows example documents from the synthetic RAG evaluation corpus (Section~\ref{sec:rag_eval}).

\subsection{Ground Truth Document}

The following internal memo contains the ground truth explanation for the changepoint:

\begin{codeboxenv}[memo\_project\_helios\_launch\_2022-07-20.txt]
INTERNAL MEMO - CONFIDENTIAL

From: Maria Chen, Chief Technology Officer
To: All Employees
Date: July 20, 2022
Subject: Project Helios Launch Success

Dear Team,

I am thrilled to announce the successful launch of Project
Helios on July 15, 2022.

On July 15, 2022, Nexora Technologies launched Project Helios,
a revolutionary AI-powered recommendation engine. The launch
resulted in a 40
uptick in monthly active users. The project was led by CTO
Maria Chen and had been in development since Q3 2021.

Key Highlights:
- User engagement increased by 40
- Monthly active users surged from 175,000 to over 210,000
- Customer satisfaction scores reached an all-time high
- The recommendation accuracy improved to 94.7%

This achievement represents 18 months of dedicated work by the
Helios team. Special thanks to the engineering leads: James
Wright, Sarah Kim, and David Okonkwo.

Best regards,
Maria Chen
CTO, Nexora Technologies
\end{codeboxenv}

\subsection{Example Distractor Document}

The following product specification is representative of the 30 distractor documents in the corpus---topically related to the company but irrelevant to the changepoint:

\begin{codeboxenv}[spec\_cloudvault\_2023-07-25.txt]
PRODUCT SPECIFICATION DOCUMENT

Product: CloudVault
Version: 3.5.3
Last Updated: 2023-07-25

Overview:
CloudVault provides enterprise-grade solutions for data
management.

Technical Requirements:
- Python 3.8+
- 8GB RAM minimum
- 100GB storage

Dependencies:
- PostgreSQL 13+
- Redis 6+
- Kubernetes 1.20+

---
Document Owner: Engineering Team
\end{codeboxenv}

\end{document}